# SHORT TERM ELECTRIC LOAD FORECAST WITH ARTIFICIAL NEURAL NETWORKS

**Vasar Cristian, Szeidert Iosif, Filip Ioan, Prostean Gabriela**

*Department of Control System Engineering, "Politehnica" University from Timisoara,
Faculty of Automation and Computer Science,
300223 Timisoara, Romania, Phone: (+40) 256-403237,
E-Mail: cristian.vasar@aut.upt.ro*

Abstract: This paper presents issues regarding short term electric load forecasting using feedforward and Elman recurrent neural networks. The study cases were developed using measured data representing electrical energy consume from Banat area. There were considered 35 different types of structure for both feedforward and recurrent network cases. For each type of neural network structure were performed many trainings and best solution was selected. The issue of forecasting the load on short term is essential in the effective energetic consume management in an open market environment. *Copyright © 2007 IFAC*

## 1. GENERAL ASPECTS

As the electrical energy stocking on a large scale is impossible, the power network's main role is to transport the consumers' load energy demand. Therefore, *it is very important to study and to analyze the evolution of the load in order to operate and to project the power network*, as all the other decisions are based on the load volume.

The load forecast is the scientific activity whose objective is the estimation of the energy and power consumption based on varied information analyses, so that we can finally obtain a concordance between the estimated consumption and the real one.

The load forecast has the following features:
- It is a dynamical activity, which is strongly influenced by the time factor;
- The correct appreciation of some uncertain factors evolution is essential for the realistic future forecast;
- In order to justify the decisions related to the power system development and operation the forecast results are strongly necessary;
- The load forecast errors imply high extra costs:
  - If the load was underestimated we have extra costs caused by the damages due to lack of energy, or by the overloading system elements;
  - If the load was overestimated, the network investment costs overtake the real needs, and the fuel stocks are overvalued, locking up, in an unjustified way, capital investment.

The main factors that affects electrical load behavior, from the experience in this field:
- *The weather conditions:* the season, the daily temperatures (medium, minimal, maximal), the wind speed, the rain-fall quantity, the cloudiness, etc.;
- *The demographic factors*: the population rate of growth, the number of the inhabitants in a certain area or in a certain country, the birth rate (the number of child bearings per 1000 inhabitants), the population growth (the difference between the birth rate and the death rate) etc.;

- *The economical factors:* the GNP, the labor productivity, the medium specific incoming, the economy development rate, the endowment level, the life quality level; a very important element is represented by the energy purchase price, this one being connected to the supply – demand ratio on a side, and to the quantity of the supplies and to the economical politics on the other side;
- *Other factors:* The length of the day compared to the length of the night influences, directly the load determined by the artificial lightening; The day of the week we are referring to, knowing that, on holidays, the load is reduced compared to the load on a work day, on which the production activity leads to a high level of load; the economical activity is more intense on the days which are in the middle of the week (on Tuesday, on Wednesday, on Thursday) than on the other days (on Monday, on Friday). (Figueiredo *et al.*, 2005)

The evolution of these parameters has a strong random character. At one time, the more or less accidental realizations of these parameters influence in a direct and univocal way the load (explained variable) and their variation tendency change influences, in a decisive way, the load variation tendency (Abdel-Aal R.E., 2004).

The load forecast implies the building of a model that uses the determinant factors, which influence the load as entry, and the corresponding consumed energy as output. This implies the following aspects:
- Identifying and classifying according to their importance the causes that influence the load;
- Determining from the qualitative and the quantitative point of view the relationship between the causes and the effect.
- Using the relationship established above for the load forecast, based on the parameters' evolution estimation.

The gradual checking, of the forecast results, as time goes on, correlating the energy consumption to its causes and accordingly modifying the forecast to the revised relationship formula in order to get better forecasts.

The issue of forecasting the load on short, medium and long term is essential in the effective energetic consume management in an open market environment. This problematic is not completely solved due to non-linear and randomness of the industrial and home consumers behavior. The electrical energy consume is also influenced by external factors such as: electrical energy price variations, weather conditions and the zonal economic growth rate. (Song *et al.,* 2006)

## 2. LOAD FORECAST USING ARTIFICIAL NEURAL NETWORKS (ANN)

One viable solution for load forecast is based on ANN, and it comes from their capacity to model complex processes, whose functioning rules are either too complicated to be algorithmically described, or too ambiguous, even unknown, to be analytically described. ANN are known for their generalization capacity, to give correct answers for the situations that are different from that of the learning process, and the synthesizing that allows to draw some conclusions when the ANN have to face special situations: incomplete, partial, or even contradictory information, strong perturbations due to noises high level (Methaprayoon, *et al.* 2006).

Elecric load forecast is performed using artificial neural networks trained using backpropragation algorithm. Data set using for training and validation was structured in different data sets, considering different delays and different number of inputs (up to 12 inputs).

There have been considered feed-forward and Elman neural networks having different number of inputs and different number of hidden neurons.. The feedforward neural network structure is presented in Figure 1 (Hippert *et al*., 1990).

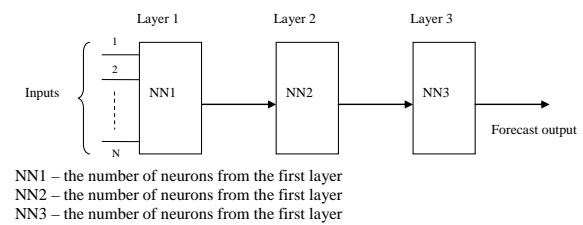

NN1 – the number of neurons from the first layer
NN2 – the number of neurons from the first layer
NN3 – the number of neurons from the first layer

Fig. 1. Feedforward neural network

There have been considered 5 neural networks based on this feedforward neural network structure. The number of neurons corresponding to each considered structure is presented in Table 1. For example network number 3 is structured in 3 layers having 5 neurons in the first layer, 7 neurons in the second and 1 neuron in the output layer. The neurons from the first and the second layer have nonlinear activation an the neurons from the output layer have linear activation function. This network can have different number of inputs, which are delayed with different time values from 1 to 4 steps.

Table 1 Implemented network structures

| Network number | Number of neurons from first layer (tansig) | Number of neurons from second layer | Number of neurons from output layer (purelin) |
|---|---|---|---|
| 1 | 3 | 3 | 1 |
| 2 | 3 | 5 | 1 |
| 3 | 5 | 7 | 1 |
| 4 | 9 | 5 | 1 |
| 5 | 12 | 10 | 1 |

The training process is conducted off-line, considering ten neural networks from each type and an initial number of 5000 epochs. The best obtained networks are trained further in order to increase the forecast precision.

Network validation is done using a supplementary set of measurements, comparing the results for a range of 12 measurements.

A similar experiment was conducted using Elman recurrent neural networks. Input data set, containing electric load measurements, was organized in different data sets considering 1, 2, 3 or 4 step delays.

In Figure 2 there is presented the Elman network structure used in this paper. Several different architectures were considered, having the number of inputs varying from 2 to 8 and the number of neurons according to Table 1.

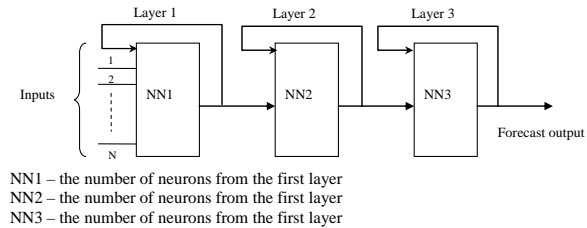

NN1 – the number of neurons from the first layer
NN2 – the number of neurons from the first layer
NN3 – the number of neurons from the first layer

Fig. 2. Elman recurrent neural network

For example the network number 1, structured in 3 layers, the first and second layer having 3 nonlinear neurons, and the output layer having 1 linear neuron, considering 2 to 8 inputs will lead to 7 different structures for this type of network. Taking into account the inputs delayed with 1, 2, 3 and 4 steps results 4 different data sets and this lead to 4 different trained networks for each considered structure.

All Elman networks were trained for a number of 5000 training epochs. The best obtained networks are trained further in order to increase the forecast precision.
After training process all the networks were validated using a testing data set in order to verify the network ability to extrapolate its knowledge beyond the data set used for training purposes.

## 3. INPUT DATA SET

The number of the data samples taken into consideration for the past, the characteristics of the used neural model and the adopted work hypothesis can lead to different realizations. They can be compared using different criteria such as:
- checking of the forecast possibilities in some known points. In order to have such checking points, the last 2-3 weeks (the latest ones) are being taken into account for the future, the forecast sphere moves towards the past, 2-3 weeks ago. The forecast results for the next 2-3 weeks can be checked, and if there is a resemblance between the actual results and the expected ones for this same period, then, it is very possible that the load increase is respected for the next years too. This has a major disadvantage, meaning that in order to establish the forecast, you do not take into account the tendencies which occurred in the latest years and which are crucial for the future evolutions.
- Comparisons using previous forecasts, forecasts done by other means, already existing data in order to estimate the obtained values.
- Heuristic approaches based on experience and on the experts' intuition

Input data is represented by the electric power consumed for a period of several months, measured every hour. The measurements were organized in different data sets used for training and respectively validation purposes, considering the sample time, the imposed delays (between two successive measurements) and the number of neural network inputs.

The sample time was considered 1 hour, and the delays were considered 1,2 3 and respectively 4 hours. For each considered delay was generated 7 data sets, corresponding to neural networks having 2,3,…8 inputs, as it is presented in Table 2, where $u(k-i)$ represents the neural network inputs (consumed electric power), and $i=1,2,..N$ (number of measurements).

Table 2 Measurements organization in different input data sets

| Delay | Neural network input number | Measurements applied to the predictor input | Output |
|---|---|---|---|
| 1 | 2 | $u(k-2),u(k-1)$ | $u(k)$ |
| 1 | 3 | $u(k-3),u(k-2),u(k-1)$ | $u(k)$ |
| 1 | … | … | … |
| 1 | 8 | $u(k-8),...,u(k-2),u(k-1)$ | $u(k)$ |
| 2 | 2 | $u(k-3),u(k-2)$ | $u(k)$ |
| 2 | … | … | … |
| 2 | 8 | $U(k-9),...,u(k-3),u(k-2)$ | $u(k)$ |
| 3 | 2 | $u(k-4),u(k-3)$ | $u(k)$ |
| 3 | … | … | … |
| 3 | 8 | $u(k-10),...,u(k-4),u(k-3)$ | $u(k)$ |
| 4 | 2 | $u(k-5),u(k-4)$ | $u(k)$ |
| 4 | … | … | … |
| 4 | 8 | $u(k-11),...,u(k-5),u(k-4)$ | $u(k)$ |

Thus 28 different data sets are obtained in order to cover many possible combinations of input data an get the data structure that improve electric load forecast error. It is obvious that the biggest data set is obtained for the minimum number of inputs and the minimum delay.

## 4. SIMULATION STUDIES

This paragraph presents the most representative results obtained for sets of ten neural networks of each type as described in Table 1. The best result after training feed-forward networks was obtained for a network with 3 layers (9 neurons on the first layer, 5 in the second and 1 in the output layer) and 7 inputs with unitary phase difference. The evolution of training error is presented in figure 3.

The training was performed on 2000 epochs but for a better remarking of error curve it is presented only the evolution on first 100 epochs. The minimum error established was $10^{-10}$ and was achieved after 92 epochs.

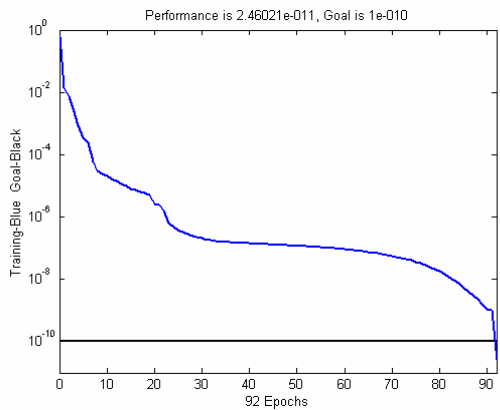

Fig. 3 Training error for 7 inputs neural network with unitary phase difference

There were performed simulation obtaining the forecast electric load for the next 3 weeks, represented comparative with the real electric load consumed (figure 4). For the training phase there was used a data set that contains the consumed energy for a 40 days time period. There can be noticed that it was obtained a quite good prediction, taking into account the reduced energy consume in the weekends, resulting unnecessary the addition of a supplemental input versus in the case of medium term prediction.

The feed forward neural networks have proved to be very good in the forecast, even if for the obtaining of the best result there was necessary a neural network with a larger number of inputs, respectively a larger amount of data samples. This fact can be justified due to the variation of the energy consume, that presents a periodicity (the time period being a week).

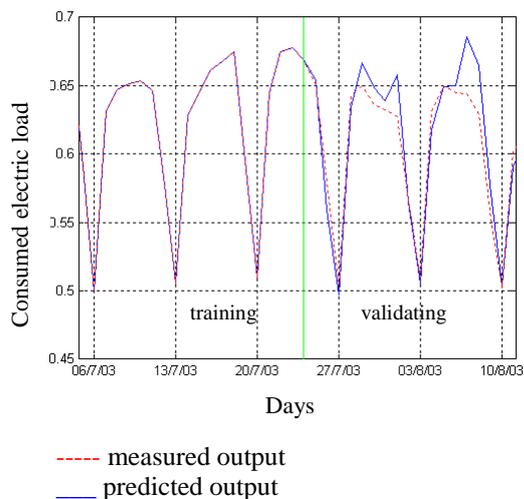

----- measured output
___ predicted output
Fig. 4 Electric load forecast for 3 weeks

For the comparative study regarding the influence of the network's inputs number on the overall performance there were chosen a data input sample with unitary phase difference and the network structure with 9, 5 and respectively 1 neuron in the layers. From the chosen structure we derived 7 neural networks having from 2 to 8 inputs. The error evolutions corresponding to the training epochs are presented in figure 5 – where in the balloons there is specified the effective number of network inputs.

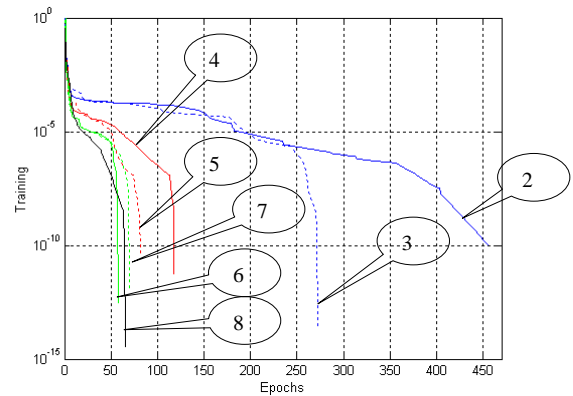

Fig. 5 Training error evolution for feedforward networks with different number of inputs

In figure 5 there can be noticed the fact that by increasing the number of the network's inputs, the error decreases. The training was performed in 2000 epochs. There is highlighted only a detail, because the error curve decreases significantly in the period of first training epochs. The aspects noticed in figure 5 reiterate the fact that for the electric energy consume forecast are required neural networks with a higher number of inputs.

The analysis of the obtained results for the training of feed forward neural networks with different structures has been performed for a 7 input neural network. The data phase difference was chosen unitary. In figure 6 there is presented the error evolution for the all five structures of neural networks (in the balloons there is specified the number of network structure according to Table 1). Due to the fast error decreasing there is highlighted only the detail for the first training epochs.

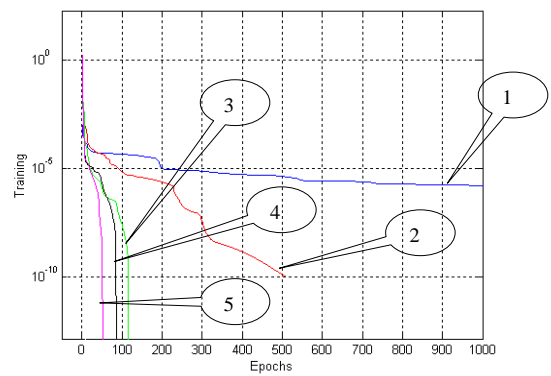

Fig. 6 Training error evolution for feedforward neural network with different number of neurons

There can be noticed the fact that by increasing the number of the network's inputs, the error decreases faster. There was considered as optimal the neural network with 9, 5, 1 neurons on the layers because the prediction error is insignificant. Another conclusion is that an increase of the neural network structure's complexity doesn't assure a corresponding increase of prediction's quality. Therefore, it can be concluded that it is not necessary nor recommended the usage of a neural network with a large complex structure for the considered application – the electrical energy consume forecast (Vasar *et al.*, 2004).

Another issue to be considered is also the influence of the input data's phase difference over the learning modality of the neural networks. This fact is important for the setup of the phase difference used in the construction of the input data set.

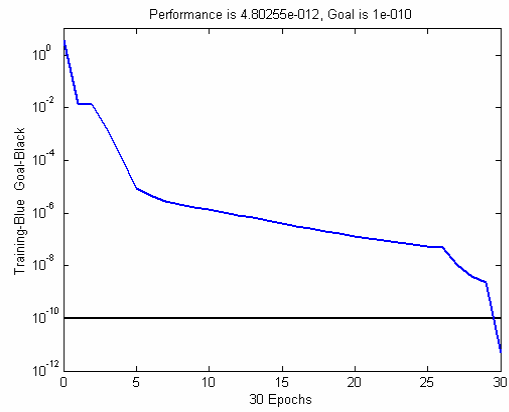

Fig. 8. Training error evolution for an Elman network with 8 inputs and 12,10,1 neurons on each layer

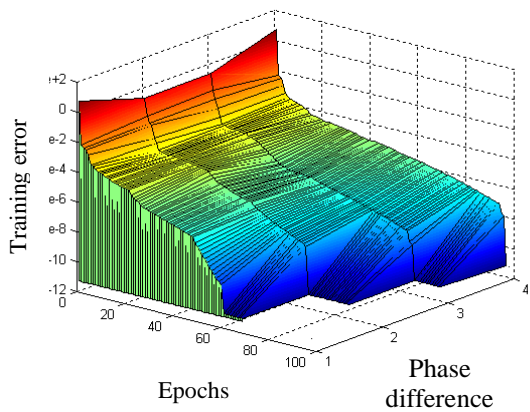

Fig. 7. Training error evolution for several phase difference

In figure 7, there can be noticed the input data set with a unitary phase difference (one day), where the difference between the prediction period and of the most recent input (phase difference) represents the best case. A one step (unitary) phase difference is the best choice for the neural network training process. There were chosen for the feed forward neural network's training a structure with 9, 5, 1 neuron on the layer and 7 inputs. In conclusion, the data constituted from the records of electrical energy consume with a unitary phase difference embed the required information for the training process of the feed forward neural networks.

The short-term prediction of the electrical energy forecast has been performed by using Elman neural networks in a similar manner to those previously stated. Below are presented only the best obtained results.

There were conducted experiments regarding the training of feed forward neural networks with the previously considered structures and the best results were obtained for the network with 3 layers having 12, 10 and respectively 1 neuron on each layer, 8 inputs, where the data samples have an unitary phase difference. (figure 8).

The training was performed in 2000 epochs with the mention that for a better highlighting of the minimum error reaching there was depicted a detail on 30 epochs. The minimum error was considered $10^{-10}$ and it was reached after 30 epochs. It can be noticed the optimal neural network has in this case a structure more complex regarding the distribution of the neurons on layers and a larger number of inputs then the one settled in the case of medium term electrical energy forecast. The phase difference step is the same as in previously case.

By using a new input data test set, there were obtained the forecasted consume for the following three weeks, represented comparatively to the effective energy consume (as depicted in figure 9). For the training process there have been used a number of 40 samples (days). There can be noticed a good forecast taking into account the extremely reduced consume in the period of weekends, there being unnecessary the addition of supplemental inputs as in the case of medium term prediction. There can be concluded that similar to the case of feed forward neural networks, for the short term forecast is necessary a higher amount of inputs in the neural network.

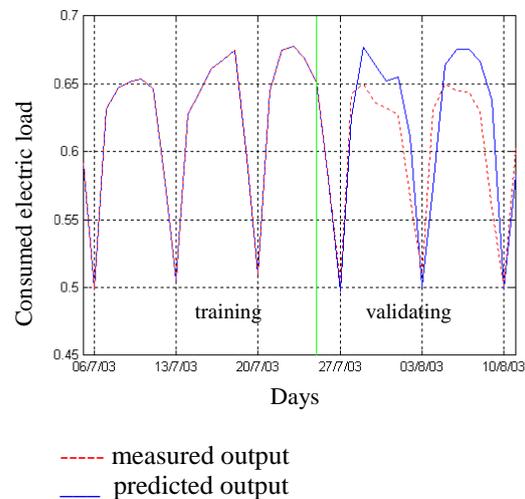

----- measured output
_____ predicted output
Fig. 9. Electric load forecast for 3 weeks

In order to study the influence of the number of network's inputs over the error evolution there has been chosen a network with 12, 10 respectively 1 neuron on layer and a unitary phase difference with several input data sets.

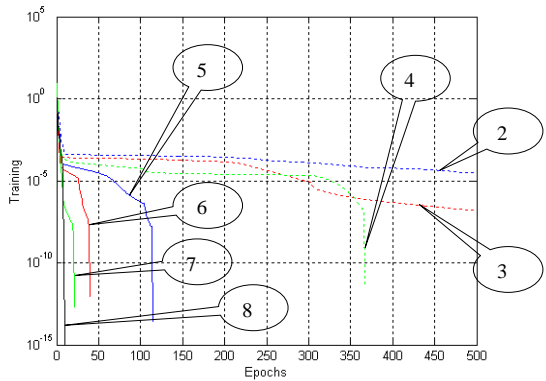

Fig. 10 Training error evolution for Elman networks with different number of inputs

In figure 10 (where in the balloons there is specified the effective number of network inputs) there can be noticed that for the networks with a small number of inputs the prediction error is unsatisfactory. The training was performed over the period of 2000 epochs. There are depicted only the first 500 epochs due to the fact that there are insignificant differences between the prediction errors of networks with large number of inputs. The analysis of the obtained results of the Elman network's training was performed considering the case of networks with 8 inputs with different numbers of neurons in the layers. The input data set was considered having a unitary phase difference, this being the case that lead to best training results. In figure 11 is presented the error evolution for the 5 types of considered neural networks (in the balloons there is specified the number of network structure according to Table 1). Due to the fast decrease of the error, there were depicted only the detail regarding the first training epochs.

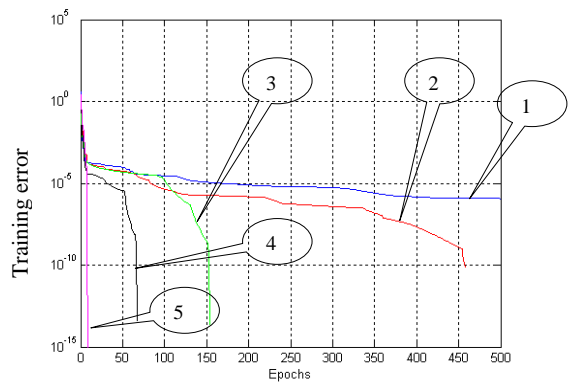

Fig. 11. Training error evolution for Elman neural network with different number of neurons

There can be noticed that all types of neural networks reach the minimum error and the more complex one, succeed to reach faster the imposed minimum error.
Another important issue is the influence of the phase difference between the prediction and the last input over the networks learning capability. This fact is capital for the setup of the phase difference step in the construction of the input data set.

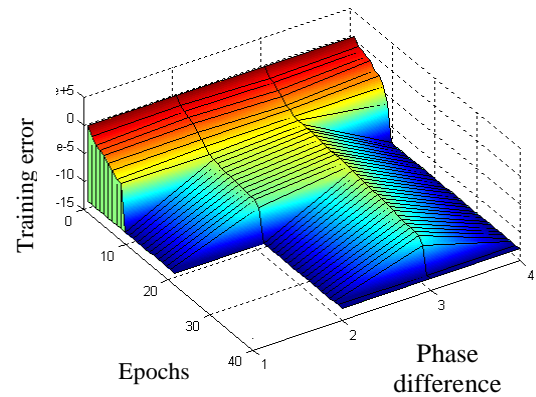

Fig. 12. Training error evolution for several phase difference

In figure 12, there can be noticed that the constructed data set with a unitary phase difference is the best choice for the neural network's training process. The Elman neural networks used for training process have 12, 10 and respectively 1 neuron on layer and 8 inputs. There can be concluded that the records of electrical energy consume considered with a unitary phase difference withhold the optimal required information for the Elman neural networks training process. Also, as in previous case, the networks with 4 step phase difference present a good error evolution in the training process.
In order to study the effects of the Elman neural networks complexity increase – through increase of number of neurons on the layer and respectively of the number of inputs there are considered the following 4 neural network's structure:
- Network no.4 – 4 inputs; structure: 3, 5, 1 neurons per layer;
- Network no. 3 – 6 inputs; structure: 5, 7, 1 neurons per layer;
- Network no. 2 – 8 inputs; structure: 9, 5, 1 neurons per layer;
- Network no. 1 – 10 inputs; structure: 12, 10, 1 neurons per layer;

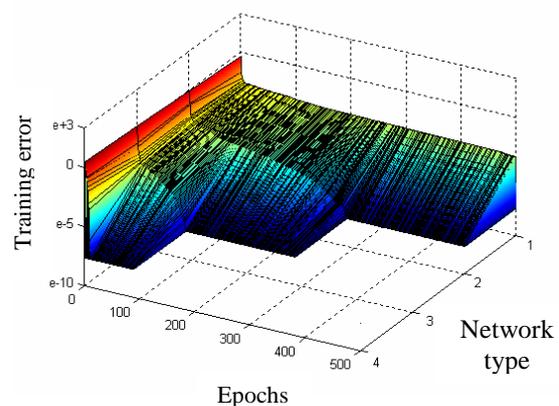

Fig. 13. Training error evolution for Elman neural network with different structures

This study presents importance from the same point of view of networks efficiency validation, being well known the fact that the one with an increased complexity are also huge calculus resources and time consumers in the training process.

There can be noticed that the usage of complex networks, having 12, 10 and respectively 1 neuron per layer and 10 inputs delivers the best results in the training process. (Figure 13). Further, the initial training error decreases with the decrease of the networks complexity.

## 5. CONCLUSION

After this study considering two types of neural networks (feedforward and Elman) for electric load short term forecasting it can be concluded that best performances were achieved using feed-forward neural networks. The Elman networks become proficient when the network structure is increased (both number of inputs and hidden neurons).

In the case of simple structures and a small number of inputs the training error of Elman's networks reaches the minimum value after a large number of training epochs, so that the training time period is quite long.

Regarding the phase difference between the moment o the last input and the prediction moment, for the Elman's networks, there has been concluded, that for each individual variable, the phase difference is one, due to fact that each individual network has prediction error of a network with different phase differences. In the most of cases the best results were obtained for unitary phase difference.

The Elman networks required a longer training period than the case of a feed-forward network, but this reflects itself in the obtained results, the results being worse. In case that there is desired the usage of Elman networks it is strongly recommended the usage of an increased number of neurons on each layer and respectively an increased number of inputs. Feed-forward networks proved a good behavior in forecasting electric load, but the training time was considerable. This is the main disadvantage of using feed-forward networks for this kind of applications.

A short term forecast of the electrical energy consumed the networks that presented the best results were the ones that presented an increased number of neurons on layers and inputs and respectively a unitary phase difference between the prediction moment and the moment of the most recent input. The obtained results were predictions for a period of 2 days and it does not guarantee that forecast is accurate for an increased time period.